\documentclass[10pt,twocolumn,letterpaper]{article}

\usepackage{iccv}
\usepackage{times}
\usepackage{epsfig}
\usepackage{graphicx}
\usepackage{amsmath}
\usepackage{amssymb}
\usepackage{subfig}
\usepackage{caption}

\usepackage{amsmath}

\newcommand{\comment}[1]{}
\newcommand{\avgunder}{\mathop{\mathrm{avg}}\limits}
\newcommand{\minunder}{\mathop{\mathrm{min}}\limits}
\newcommand{\norm}[1]{\left\lVert#1\right\rVert}

\usepackage{enumitem}
\usepackage{booktabs}
\usepackage{multirow}
\usepackage{graphicx}
\usepackage{dblfloatfix}

\usepackage[symbol*]{footmisc}
\usepackage{pifont}
\usepackage{authblk}
\usepackage[export]{adjustbox}
\newcommand{\cmark}{\ding{61}}%
\newcommand{\zmark}{\ding{67}}%

\usepackage[breaklinks=true,bookmarks=false]{hyperref}

\iccvfinalcopy 

\def\iccvPaperID{3375} 

\ificcvfinal\fi
\begin{document}

\graphicspath {{figures/}}

\title{DPOD: 6D Pose Object Detector and Refiner}

\author[,\zmark,\cmark]{Sergey Zakharov \thanks{These authors contributed equally to the work.}}
\newcommand\CoAuthorMark{\footnotemark[1]} 
\author[,\zmark,\cmark]{Ivan Shugurov $^*$} 
\author[\zmark,\cmark]{Slobodan Ilic}
\affil[\zmark]{ Technical University of Munich \quad\quad\quad  \cmark~Siemens Corporate Technology
{\normalsize \tt{sergey.zakharov@tum.de}}, {\normalsize \tt{\, ivan.shugurov@tum.de}}, {\normalsize \tt{\, slobodan.ilic@siemens.com}}}

\maketitle
\ificcvfinal\thispagestyle{empty}\fi

\begin{abstract}

In this paper we present a novel deep learning method for 3D object detection and 6D pose estimation from RGB images. Our method, named DPOD (Dense Pose Object Detector), estimates dense multi-class 2D-3D correspondence maps between an input image and available 3D models. Given the correspondences, a 6DoF pose is computed via PnP and RANSAC. An additional RGB pose refinement of the initial pose estimates is performed using a custom deep learning-based refinement scheme. Our results and comparison to a vast number of related works demonstrate that a large number of correspondences is beneficial for obtaining high-quality 6D poses both before and after refinement.
Unlike other methods that mainly use real data for training and do not train on synthetic renderings, we perform evaluation on both synthetic and real training data demonstrating superior results before and after refinement when compared to all recent detectors. While being precise, the presented approach is still real-time capable.

\end{abstract}

\section{Introduction}

Object detection has always been an important problem in computer vision and a large body of research has been dedicated to it in the past. This problem, like many other vision problems, witnessed a complete renaissance with the advent of deep learning. Detectors like R-CNN~\cite{girshickRichFeatureHierarchies2014a}, and its follow-ups Fast-RCNN~\cite{girshickFastRcnn2015}, Faster-RCNN~\cite{renFasterRCNNRealTime2015}, Mask-RCNN~\cite{heMaskRCNN2017}, then YOLO~\cite{redmonYouOnlyLook2015} and SSD~\cite{liuSSDSingleShot2016} marked this research field with excellent performance. All these works localize objects of interest in images in terms of tight bounding boxes around them. However, in many applications, e.g., augmented reality, robotics, machine vision, etc., this is not enough and a full 6D pose is necessary. While this problem is easier to solve in depth images, in RGB images it is still quite challenging due to perspective ambiguities and significant appearance changes of the object when seen from different viewpoints. 


\begin{figure}[t]
	\centering
	\includegraphics[width=0.99\linewidth]{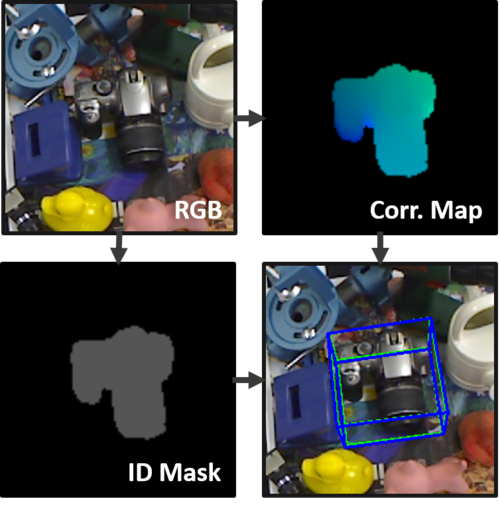}
	\caption{\textbf{Example output of the DPOD method:} Given
a single RGB image, we regress its ID mask and its 2D-3D correspondences. PnP+RANSAC is then applied to estimate the final pose. The green bounding box shows the ground truth pose, while the blue one corresponds to the
estimated pose. The almost perfect overlap of the bounding boxes
indicates that estimations are very accurate.
 \label{figure:into_vis}}
\end{figure}

Recent deep learning-based approaches, such as SSD6D~\cite{kehlSSD6DMakingRGBbased2017b}, YOLO6D~\cite{tekinRealTimeSeamlessSingle2017}, AAE~\cite{sundermeyerImplicit3DOrientation2018}, PoseCNN~\cite{YuXiang17_PoseCNN} and PVNet~\cite{SPeng18_PVNet}, are the current top performers for this task in RGB images. Even though they all perform evaluation on LineMOD and OCCLUSION datasets, each of them focuses on different aspects of the 6D pose estimation pipeline. The majority is trained on real data~\cite{tekinRealTimeSeamlessSingle2017,YuXiang17_PoseCNN,SPeng18_PVNet,OJafari17_iPose} while only SSD6D~\cite{kehlSSD6DMakingRGBbased2017b} and AAE~\cite{sundermeyerImplicit3DOrientation2018} are trained on synthetic renderings. Some are presented without refinement, like YOLO6D~\cite{tekinRealTimeSeamlessSingle2017} and PoseCNN~\cite{YuXiang17_PoseCNN}, while the others perform refinement. The most recent refiners are based on deep learning, e.g., DeepIM~\cite{LiWJXF18_DeepIM} that acts on poses from the PoseCNN detector and the refiner of Manhardt et al.~\cite{ManhardtKNT18} that uses SSD6D poses. 


Inspired by the methods of Gueler et al.~\cite{gulerDenseposeDenseHuman2018} and Taylor et al.~\cite{taylor2012vitruvian}, which estimate dense correspondences between the human body model and the humans in the image, we propose a novel 3D object detector and pose estimator that also estimates dense 2D-3D correspondences. Unlike DensePose for humans, which requires a sophisticated annotation tool and enormous annotation efforts, our method is annotation-free and only requires creation of arbitrary UV texture maps of the objects, that we do automatically---mainly by spherical projections. The two key elements of our approach are: the pixel-wise prediction of the multi-class object ID masks and classification of correspondence maps that directly provide a relation between image pixels and 3D model vertices. In this way, we end up with a large number of pixel-wise correspondences, which allow for a much better pose estimation than, for example, 9 regressed virtual points of the object's bounding box as in YOLO6D. 

In addition to this, we introduce a deep learning-based pose refinement network that takes initial poses estimated with our DPOD detector and enhances them. The proposed refinement approach builds on the successes of \cite{LiWJXF18_DeepIM,manhardtDeepModelBased6D}, but is shown to be faster, simpler to train, able to be trained both on synthetic and real data, and it outperforms the former solutions in terms of pose quality. We demonstrate that 
even our poses, which are already of high quality, can be further improved with our refiner.

We experimented by training our detector with only synthetic and only real images. In both cases, our unified method, named DPOD, composed of the dense pose detector and the refiner outperforms other related works. Dense correspondences not only allow for standard PnP and RANSAC to estimate accurate poses without refinement, but also pave the way for a successful pose refinement. For the models trained on real data, one iteration of refinement is enough to outperform all other reported results, even SSD6D with the depth-based ICP refinement.


In the remainder of the paper we first review related approaches, then introduce our approach, explaining data preparation, training, architectures and  pose refinement. Finally, we present an exhaustive experimental validation and comparison with recent works, where we demonstrate the superiority of our approach.

\section{Related Work}

Detecting 3D objects and estimating their 6D pose has been addressed in many works in the past, but the majority of them used depth or RGB-D cameras~\cite{brachmannLearning6DObject2014a,choiRGBDObjectPose2016, kehlDeepLearningLocal2016a,laiScalableTreebasedApproach2011, michelGlobalHypothesisGeneration2017,sockMultiview6DObject2017,caoCombinedHolisticLocal2017}. Depth information disambiguates the object's scale that is the most critical in RGB images due to perspective projection. Therefore, using only RGB images for detection and 6D pose estimation is a quite challenging problem. Recent solutions are mainly based on deep learning and automatically learned features, while older ones use hand-crafted features or image information, e.g., gradients, or image pixel intensities directly.

Template matching approaches, e.g., \cite{hinterstoisserMultimodalTemplatesRealtime2011a, hinterstoisserModelBasedTraining2013, rios-cabreraDiscriminativelyTrainedTemplates2013}, render synthetic image patches from different viewpoints distributed on a sphere around the 3D model of the object and store them as a database of templates. Then the input images are searched using this template database sequentially in a sliding window fashion. Efficient and robust template matching strategies have been presented for color, depth and RGB-D images. The most popular approach is arguably LineMOD~\cite{hinterstoisserMultimodalTemplatesRealtime2011a}, which also provided a first dataset with labeled poses. This dataset is still used as a benchmark for object detection and pose estimation.  Another alternative to template matching approaches is the learning approaches that employ random forests~\cite{brachmannLearning6DObject2014a, brachmannUncertaintyDriven6DPose2016a, cavallari2019real}.

\vspace{-1em}
\paragraph{Deep Learning 6D Pose Detectors.}
In the last two years deep learning approaches have shown that impressive results can be obtained for detection and pose estimation in RGB images. 
Here we review the following ones: SSD6D~\cite{kehlSSD6DMakingRGBbased2017b}, YOLO6D~\cite{tekinRealTimeSeamlessSingle2017}, BB8~\cite{radBB8ScalableAccurate2017b}, iPose~\cite{OJafari17_iPose}, AAE~\cite{sundermeyerImplicit3DOrientation2018}, PoseCNN~\cite{YuXiang17_PoseCNN} and PVNet~\cite{SPeng18_PVNet}.

SSD6D~\cite{kehlSSD6DMakingRGBbased2017b} extended the ideas of the 2D object detector~\cite{liuSSDSingleShot2016} by 6D pose estimation based on a discrete viewpoint classification rather than direct regression of rotations. 
The method is rather slow and poses predicted this way are quite inaccurate since they are only a rough discrete approximation of the real poses. The refinement is a must in order to produce presentable results. BB8~\cite{radBB8ScalableAccurate2017b} uses a three-stage approach. In the first two stages the coarse-to-fine segmentation is performed, the result of which is then fed to the third network trained to output projections of the object's bounding box points. Knowing 2D-3D correspondences, a 6D pose can be estimated with PnP. The main disadvantage of this pipeline is its multi-stage nature, resulting in very slow run times. 
Building on YOLO and BB8 ideas, YOLO6D~\cite{tekinRealTimeSeamlessSingle2017} proposed a novel deep learning architecture capable of efficient and precise object detection and pose estimation without refinement. As is the case with BB8, the key feature here is to perform the regression of reprojected bounding box corners in the image. The advantages of this parametrization are its relative compactness and that it does not introduce a pose ambiguity as opposed to a direct regression of the rotation. Moreover, in contrast to SSD6D, it does not suffer from pose discretization resulting in much more accurate pose estimates without refinement. 

Among the methods that are specifically designed to be robust to occlusions we would like to highlight iPose~\cite{OJafari17_iPose}, PoseCNN~\cite{YuXiang17_PoseCNN}, and PVNet~\cite{SPeng18_PVNet}.
iPose~\cite{OJafari17_iPose} operates in 3 separate stages: segmentation, 3D coordinate regression and pose estimation. By contrast, our approach unifies the first two stages into the end-to-end network. Moreover, we do not regress 3D coordinates, but rather UV maps that turned out to be a much easier task for the network, resulting in less erroneous correspondences. PoseCNN~\cite{YuXiang17_PoseCNN} also estimates object masks, but then separately estimates the translation of the object's centroid and regresses a quaternion for rotation. PVNet~\cite{SPeng18_PVNet} takes a different approach and designs a network which for every pixel in the image regresses an offset to some predefined keypoints. Instead of bounding box points, they vote for the points located on the object itself. This allows them to handle occlusions very well. 
AAE (Augmented Autoencoders)~\cite{sundermeyerImplicit3DOrientation2018} concentrates on pose estimation and training from synthetic models, while using already computed SSD detection bounding boxes as input.

\vspace{-1em}
\paragraph{Deep Learning 6D Pose Refiners.}
Deep learning-based 6D pose refinement has shown promising results in recent publications \cite{manhardtDeepModelBased6D, LiWJXF18_DeepIM}. Both refiners are conceptually very similar and are designed to output relative transformation between the real input image patch and the patch containing the object rendered with the predicted pose. Main differences are the used backbone architectures and loss functions. Both refinement algorithms rely on external object detection and pose estimation algorithms: for DeepIM~\cite{LiWJXF18_DeepIM} it is PoseCNN, for \cite{manhardtDeepModelBased6D} it is SSD6D \cite{kehlSSD6DMakingRGBbased2017b}. The former relies on real data, whereas the latter focuses on training on synthetic images. We propose a network architecture which takes the best of above architectures and is independent of the type of training data used.

Our work differs from the above approaches by being a complete end-to-end pipeline integrating a detector and pose estimator based on dense correspondences. We demonstrate that we can train either from real or synthetic data and in both cases we outperform all related approaches by a large margin on the LineMOD and OCCLUSION datasets.

\section{Methodology}

In this section we first discuss the training data preparation steps, followed by the neural network architecture and loss functions used, as well as the pose estimation step from dense correspondences. Finally, we describe our deep learning model-based pose refiner.  

\begin{figure*}[t]
	\centering
	\includegraphics[width=1\linewidth]{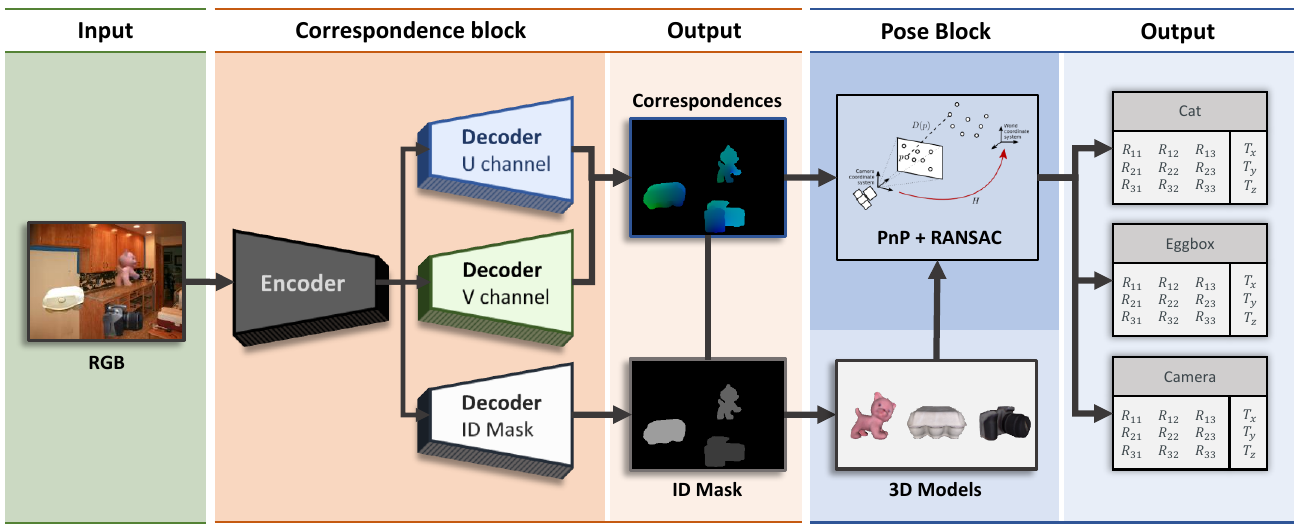}
	\caption{\textbf{Pipeline description:}  Given an input RGB image, the correspondence block, featuring an encoder-decoder neural network, regresses the object ID mask and the correspondence map. The latter one provides us with explicit 2D-3D correspondences, whereas the ID mask estimates which correspondences should be taken for each detected object. The respective 6D poses are then efficiently computed by the pose block based on PnP+RANSAC.}
	\label{fig:pipeline}  
\end{figure*}

\subsection{Data Preparation}

Most recent RGB-based detectors can be divided in two groups based on the type of data they use for training: synthetic-based and real-based. The first group of methods, e.g., SSD6D~\cite{kehlSSD6DMakingRGBbased2017b} and AAE~\cite{sundermeyerImplicit3DOrientation2018}, makes use of textured 3D models, usually provided with the public 6D pose detection datasets. The objects are rendered from different viewpoints, producing a synthetic training set. The methods of the second group on the other hand, e.g., BB8~\cite{radBB8ScalableAccurate2017b}, YOLO6D~\cite{tekinRealTimeSeamlessSingle2017}, PVNet~\cite{SPeng18_PVNet}, use the training split of the real dataset. They utilize ground truth poses provided with the dataset and compute object masks to crop the objects from real images producing a training set. 

Both types of data generation have their pros and cons. When real images sufficiently covering the object are available, it is more advantageous to use them for training. The reason is that their close resemblance to the actual objects allows for faster convergence and better results. However, training on real images biases the detector to light conditions, poses, scales and occlusions present in the training set, which might lead to problems with generalization in new environments. When, however, no pose annotations are available, which can often be the case since acquiring pose annotations is an expensive process, we are left with 3D models of the objects. With synthetic renderings, one can produce a virtually infinite number of images from different viewpoints. Despite being advantageous in terms of the pose coverage, one has to deal with the domain gap problem severely hindering the performance if no additional data augmentation is applied. Potentially, one can benefit from the advantages of both data types by mixing real and synthetic data in the training set. Therefore, approaches which can be trained on both types of data are desirable. Since our pipeline is not data-specific, we show how to generate the training data for both scenarios.



\vspace{-1em}
\paragraph{Synthetic Training Data Generation.} Given 3D models of the objects of interest, the first step is to render them from different poses sufficiently covering the object. The poses are sampled from the half-sphere above the object. Additionally, in-plane rotations of the camera around its viewing direction from -30 to 30 degrees are added. Then, for each of the camera poses, an object is rendered on a black background and both RGB and depth channels are stored. 

Having the renderings at hand, we use a generated depth map as a mask to define a tight bounding box for each generated rendering. Cropping the image with this bounding box position, we store RGB patches, masks separating them from the background, and the camera poses. At this point, we have everything ready for the online augmentation stage, which is described in the later subsection. This step of data preparation is identical for the detector and for the refinement pipelines.

\vspace{-1em}
\paragraph{Real Training Data Generation.}
In this case, an available dataset with pose annotations is divided into non-overlapping train and test subsets. Here, we follow the protocol defined by BB8~\cite{radBB8ScalableAccurate2017b} and YOLO6D~\cite{tekinRealTimeSeamlessSingle2017} and use 15\% of data for training and the rest 85\% for evaluation. Poses are selected such that the relative orientation between them is larger than a certain threshold. This approach guarantees that selected poses cover the object from all sides. For training the detector, objects are cut out from the original image using the provided mask and then stored as patches for the online augmentation stage. Additional in-plane rotations are added to artificially simulate new poses. For training the refinement, objects are left as they are.

\begin{figure}[b]
  \centering
  \includegraphics[width=0.95\linewidth]{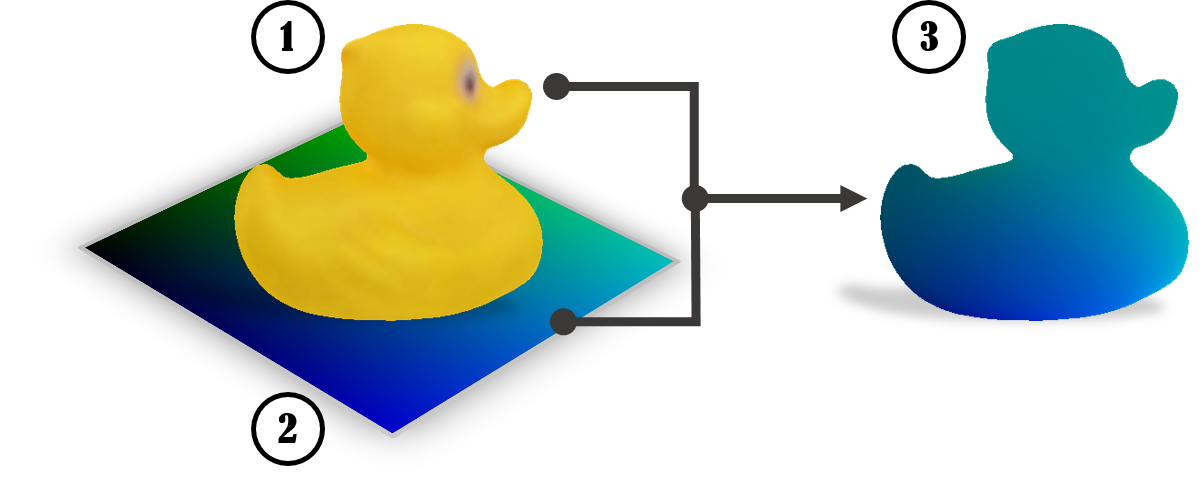}
  \caption{
  \textbf{Correspondence model:} Given a 3D model of interest (1), we apply a 2 channel correspondence texture (2) to it. The resulting correspondence model (3) is then used to generate GT maps and estimate poses.
  }

  \label{fig:uv_map}
\end{figure}

\subsubsection{Correspondence Mapping}
To be able to learn dense 2D-3D correspondences, each model of the dataset is textured with a correspondence map (see Figure~\ref{fig:uv_map}). A correspondence map is a 2-channel image with values ranging from 0 to 255. Objects are textured using either simple spherical or cylindrical projections.
Once textured, we get a bijective mapping between the model's vertices and pixels on the correspondence map. This provides us with easy-to-read 2D-3D correspondences since given the pixel color, we can instantaneously estimate its position on the model surface by selecting the vertex with the same color value. For convenience, we call the copies of the original models textured with the correspondence map {\em correspondence models}. Given the predicted correspondence map, we estimate the object pose with respect to the camera using the pose estimation block, which is described later. Similar to the synthetic or real data generation steps, we render correspondence models under the same poses as for training data and store correspondence patches for each RGB patch.


\subsubsection{Online Data Generation and Augmentation}
\paragraph{Detection and Pose Estimation.} The final stage of data preparation is the online data generation pipeline, which is responsible for providing full-sized RGB images ready for training. Generated patches (real or synthetic) are rendered on top of images from MS COCO dataset~\cite{linMicrosoftCOCOCommon2014a} producing training images containing multiple objects. It is an important step, which ensures that the detector generalizes to different backgrounds and prevents it from overfitting to backgrounds seen during training. Moreover, it forces the network to learn the model's features needed for pose estimation rather than to learn contextual features which might not be present in images when the scene changes. This step is performed no matter whether the training is being done with synthetic or real patches. We additionally augment the RGB image by random changes in brightness, saturation, and contrast, and by adding Gaussian noise. Moreover, object ID masks and correspondence patches are also rendered on top of the black background in order to generate ground truth correspondence maps. An object ID mask is constructed by assigning a class ID number to each pixel that belongs to the object.

\vspace{-1em}
\paragraph{Pose Refinement.} In the case of pose refinement, pairs of images containing the object in the current (searched) pose and in the predicted pose are provided to the network. The final stage of data preparation differs considerably depending on the type of data used. In case of synthetic data, images are generated by in-painting objects on random backgrounds in a current pose. A crucial part of the augmentation is to add random light sources for every image. If real images are used for training, no in-painting is performed. In any case, produced images are further augmented as discussed above. Then a random pose is sampled around the current pose simulating the predicted pose from the detector, which will be used as an original guess of the poses to be refined. It is crucial to choose the proper prior distribution from which distorted poses are sampled.

\section{Dense Object Detection Pipeline}
Our inference pipeline is divided into two blocks: the correspondence block and the pose block (see Figure~\ref{fig:pipeline}). In this section, we provide their detailed description.

\vspace{-1em}
\paragraph{Correspondence Block.} The correspondence block consists of an encoder-decoder convolutional neural network with three decoder heads which regress the ID mask and dense 2D-3D correspondence map from an RGB image of size 320$\times$240$\times$3. The encoder part is based on a 12-layer ResNet-like~\cite{heDeepResidualLearning2016} architecture featuring residual layers that allow for faster convergence. The decoders upsample the feature up to its original size using a stack of bilinear interpolations followed by convolutional layers. However, in principle the proposed method is agnostic to a particular choice of encoder-decoder architecture. Any other backbone architectures can be used without any need to change the conceptual principles of the method. For the ID mask head the output is a $H$$\times$$W$$\times$$O$ tensor, where $H$ and $W$ are the height and width of the original input image and $O$ is the number of objects in the dataset plus one additional class for background. Similar to the ID mask head, the two correspondence heads regress tensors with the following dimensions $H$$\times$$W$$\times$$C$, where $C$ stands for the number of unique colors of the correspondence map, i.e., 256. Each channel of the output tensors stores the probability values for the class corresponding to the channel number. Once tensors are regressed, we store them as single channel images where each pixel stores the class with the maximal estimated probability, forming the ID mask, U and V channels of the correspondence image.

Formulating color regression problem as discrete color class classification problem proved to be useful for much faster convergence and for the superior quality of 2D-3D matches. Initial experiments on direct coordinate regression showed very poor results in terms of correspondence quality. The main reason for the problem was the infinite continuous solution space, i.e., $[-1; 1]^3$, where $3$ is the number of dimensions and $[-1, 1]$ is the normalized coordinate range of a 3D model. Classification of the discretized 2D correspondences allowed for a huge boost of the output quality by dramatically decreasing the output space (now $256^2$, where $256$ is the size of a single UV map dimension). Moreover, this parametrization also ensures that 3D points of the predicted correspondences always lie on the object surface. 

The network parameters are optimized subject to the composite loss function:
\begin{equation}
\mathcal{L} = \alpha \mathcal{L}_m + \beta \mathcal{L}_u + \gamma \mathcal{L}_v,
\end{equation}
where $\mathcal{L}_m$ is the mask loss, and $\mathcal{L}_u$ and $\mathcal{L}_v$ are the losses responsible for the quality of the U and V channels of the correspondence image. $\alpha, \beta,$ and $\gamma$ are weight factors set to $1$ in our case. Both $\mathcal{L}_u$ and $\mathcal{L}_v$ losses are defined as multi-class cross-entropy functions, whereas $\mathcal{L}_m$ uses the weighted version of it.





\vspace{-1em}
\paragraph{Pose Block.}
The pose block is responsible for the pose prediction. Given the estimated ID mask, we can observe which objects were detected in the image and their 2D locations, whereas the correspondence map maps each 2D point to a coordinate on an actual 3D model. The 6D pose is then estimated using the Perspective-n-Point (PnP)~\cite{zhangFlexibleNewTechnique2000} pose estimation method that estimates the camera pose given correspondences and intrinsic parameters of the camera. Since we get a large set of correspondences for each model, RANSAC is used in conjunction with PnP to make camera pose prediction more robust to possible outliers. For the results presented in the evaluation section, for each pose we run 150 RANSAC iterations with the reprojection error threshold set to $1$. 

\section{Deep model-based pose refinement}

The proposed pose refiner is a natural extension of refiners presented in \cite{manhardtDeepModelBased6D, LiWJXF18_DeepIM} and relies on the strengths of both approaches. Similar to~\cite{manhardtDeepModelBased6D, kehlSSD6DMakingRGBbased2017b, hinterstoisserPreTrainedImageFeatures2017a} we exploit an idea of using a network already pre-trained on ImageNet as a backbone architecture. Analogous to the detector, we used a ResNet-based architecture. Similar to \cite{LiWJXF18_DeepIM}, our loss function for pose estimation is the ADD measure with a more robust $L_1$ norm:
\begin{equation}
\label{add_standard}
m = \avgunder_{\mathbf{x} \in \mathcal{M}_s} \norm{ (\mathbf{Rx} + \mathbf{t}) - (\mathbf{\hat{R}x} + \mathbf{\hat{t}}) }_1,
\end{equation}
representing the vertex to vertex distance between the object in a ground truth pose and predicted pose. $\mathbf{R}$, $\mathbf{t}$ denote the ground truth pose rotation and translation, whereas $\mathbf{\hat{R}}$ and $\mathbf{\hat{t}}$ denote the predicted transformation; $\mathcal{M}_s$ is a set of points sampled from the CAD model. Points are resampled at every iteration. The number of sampled points was limited to ten thousand in order to ensure the efficiency of training iterations and reasonable memory consumption.

\begin{figure}[b]
  \centering
  \includegraphics[width=1\linewidth]{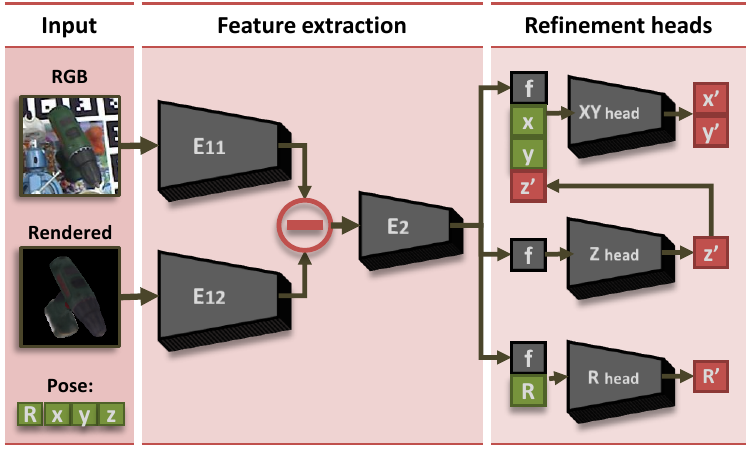}
  \caption{
  \textbf{Refinement architecture:} The network predicts a refined pose given an initial pose proposal. Crops of the real image and the rendering are fed into two parallel branches. The difference of the computed feature tensors is used to estimate the refined pose.\label{fig:refinement}
  }
\end{figure}

In Figure~\ref{fig:refinement} we show a schematic representation of the refiner. In order to be able to benefit from the network weights pretrained on ImageNet, the network has two parallel input branches, each composed of the first five ResNet layers. These layers are initialized from the pre-trained network. One branch receives an input image patch ($E_{11}$), while the other ($E_{12}$) one extracts features from the rendering of the object in the predicted pose. Then features $\mathbf f_r$ and $\mathbf f_s$ from these two networks are subtracted and fed into the next ResNet block ($E_2$) producing the feature vector $\mathbf f$. If the refinement is trained on synthetic data, it is essential to keep the first five layers unchanged and use them as the feature extractor as was shown in \cite{liuSSDSingleShot2016,hinterstoisserPreTrainedImageFeatures2017a,manhardtDeepModelBased6D}. Freezing the branch that extracts features from object renderings is unnecessary as it always operates on synthetic data. The network ends with three separate output heads: one for regressing the rotation, one for regressing the translation in X and Y directions, and one for regressing the translation in Z direction. We opted for three separate heads as the scale of their outputs is different. 
Each head is implemented as two fully connected layers. 

Rotation is always represented in the object coordinate system, which ensures that identically looking objects have the same rotation and that the network does not have to learn a more complicated transformation which arises if the world coordinate system is used. The first layer of the rotation regression head takes the feature vector $\mathbf f$ produced by ResNet and adds four values, which are the quaternion representing an initial rotation. The second layer takes the output of the previous one, stacks with the initial quaternion and outputs the final rotation. 

The head responsible for the regression of X and Y translations operates in the coordinate system of the image rather than in the full 3D space, which significantly restricts the space of possible solutions. Similar to the rotation head, the XY regression head takes the initial 2D location of the object as input and refines it. Additionally, it takes the refined prediction of Z translation. 

Weights of the fully connected layers are initialized in such a way that for the 0th iteration the network just outputs the input pose, and then during training learns how to refine those values. That significantly increases stability and speed of the training procedure as the network produces meaningful results from the very start. 

\section{Training Details}
Our pipeline is implemented using the Pytorch deep learning framework. All the experiments were conducted on an Intel Core i7-6900K CPU 3.20GHz with NVIDIA TITAN X (Pascal) GPU.
To train our method, we used the ADAM solver with a constant learning rate of $3 \times 10^{-4}$ and weight decay of $3 \times 10^{-5}$.

When training on synthetic data, the problem of domain adaptation becomes one of the main challenges. Training the network without any prior parameter initialization makes it impossible to generalize to the real data. The easy solution to this problem was proposed in several works, including \cite{hinterstoisserPreTrainedImageFeatures2017a, manhardtDeepModelBased6D}, where they freeze the first layers of the network trained on a large dataset of real images, e.g., ImageNet~\cite{ImageNetLargescaleHierarchical} or MS COCO~\cite{linMicrosoftCOCOCommon2014a}, for the object classification task.  The common observation that the authors conclude is that these layers, learning low-level features, very quickly overfit to the perfect object renderings. We follow this setup, and freeze the first five layers of our encoder initialized with the weights of the same network pretrained on ImageNet. Last but not least, we found it crucial for the performance of the detector to use various light sources during the rendering of synthetic views to account for changing light conditions and shadows in the real data.

\setlength{\tabcolsep}{6pt}
\begin{table*}[!t]
  \centering
  \caption{\textbf{Pose estimation performance:} Comparison of our approach to the other RGB detectors on the LineMOD dataset. The table reports the percentages of correctly estimated poses w.r.t. the ADD score. Among the methods trained on synthetic data, our method shows the best results significantly surpassing the former state-of-the-art. The variant of our method trained on real data again demonstrates outstanding performance outperforming most of the competitors. Moreover, our new refinement pipeline improves the estimated poses even further and shows the best overall results.}
  \label{tab:add}
  \resizebox{1\textwidth}{!}{%
    \begin{tabular}{r|ccc|cc|cccc|cc}
    \toprule
    \multicolumn{1}{r|}{\textbf{Train data}} & \multicolumn{3}{c|}{\textbf{Synthetic}} & \multicolumn{2}{c|}{\textbf{+ Refinement}} & \multicolumn{4}{c|}{\textbf{Real}} & \multicolumn{2}{c}{\textbf{+ Refinement}} \\
    \midrule
    \textbf{Object} & \textbf{SSD6D}~\cite{kehlSSD6DMakingRGBbased2017b} & \textbf{AAE}~\cite{sundermeyerImplicit3DOrientation2018} & \textbf{Ours} & \textbf{SSD6D}~\cite{manhardtDeepModelBased6D} & \textbf{Ours} & \textbf{YOLO6D}~\cite{tekinRealTimeSeamlessSingle2017} & \textbf{PoseCNN}~\cite{YuXiang17_PoseCNN} & \textbf{PVNet}~\cite{SPeng18_PVNet} & \textbf{Ours} & \textbf{DeepIM}~\cite{LiWJXF18_DeepIM} & \textbf{Ours} \\
    \midrule
    Ape & 2.6 & 3.96 & \textbf{37.22} & - & 55.23 & 21.62 & - & 43.62 & \textbf{53.28} & 77.0 & \textbf{87.73} \\
    Benchvise & 15.1 & 20.92 & \textbf{66.76} & - & 72.69 & 81.80 & - & \textbf{99.90} & 95.34 & 97.5 & \textbf{98.45} \\
    Cam & 6.1 & \textbf{30.47} & 24.22 & - & 34.76 & 36.57 & - & 86.86 & \textbf{90.36} & 93.5 & \textbf{96.07} \\
    Can & 27.3 & 35.87 & \textbf{52.57} & - & 83.59 & 68.80 & - & \textbf{95.47} & 94.10 & 96.5 & \textbf{99.71} \\
    Cat & 9.3 & 17.90 & \textbf{32.36} & - & 65.10 & 41.82 & - & \textbf{79.34} & 60.38 & 82.1 & \textbf{94.71} \\
    Driller & 12.0 & 23.99 & \textbf{66.60} & - & 73.32 & 63.51 & - & 96.43 & \textbf{97.72} & 95.0 & \textbf{98.80} \\
    Duck & 1.3 & 4.86 & \textbf{26.12} & - & 50.04 & 27.23 & - & 52.58 & \textbf{66.01} & 77.7 & \textbf{86.29} \\
    Eggbox & 2.8 & \textbf{81.01} & 73.35 & - & 89.05 & 69.58 & - & 99.15 & \textbf{99.72} & 97.1 & \textbf{99.91} \\
    Glue & 3.4 & 45.49 & \textbf{74.96} & - & 84.37 & 80.02 & - & \textbf{95.66} & 93.83 & \textbf{99.4} & 96.82 \\
    Holepuncher & 3.1 & 17.60 & \textbf{24.50} & - & 35.35 & 42.63 & - & \textbf{81.92} & 65.83 & 52.8 & \textbf{86.87} \\
    Iron & 14.6 & 32.03 & \textbf{85.02} & - & 98.78 & 74.97 & - & 98.88 & \textbf{99.80} & 98.3 & \textbf{100} \\
    Lamp & 11.4 & \textbf{60.47} & 57.26 & - & 74.27 & 71.11 & - & \textbf{99.33} & 88.11 & \textbf{97.5} & 96.84 \\
    Phone & 9.7 & \textbf{33.79} & 29.08 & - & 46.98 & 47.74 & - & \textbf{92.41} & 74.24 & 87.7 & \textbf{94.69} \\
    \midrule
    Mean & 9.1 & 28.65 & \textbf{50} & 34.1 & \textbf{66.43} & 55.95 & 62.7 & \textbf{86.27} & 82.98 & 88.6 & \textbf{95.15} \\
    \bottomrule
    \end{tabular}%
    }
  \label{tab:addlabel}%
\end{table*}%

\section{Evaluation}
In this section we evaluate our algorithm in terms of its pose and detection performance, as well as its runtime, and compare it with the state of the art RGB detector solutions. 

\subsection{Datasets}
All experiments were conducted on LineMOD \cite{hinterstoisserModelBasedTraining2013} and OCCLUSION \cite{brachmannLearning6DObject2014a} datasets, as they are the standard datasets for evaluation of object detection and pose estimation methods. The LineMOD dataset consists of $13$ sequences, each containing ground truth poses for a single object of interest in a cluttered environment. CAD models for all the objects are provided as well. The OCCLUSION dataset is an extension of LineMOD, suitable for testing how well detectors can deal with occlusions. Although it comprises only one sequence, all visible objects from the LineMOD dataset are supplied with their poses.

\subsection{Evaluation Metrics}

We evaluate the quality of 6DoF pose estimation following the procedure suggested at SSD6D \cite{kehlSSD6DMakingRGBbased2017b} also used in other papers. Analogously to other related papers \cite{tekinRealTimeSeamlessSingle2017,kehlSSD6DMakingRGBbased2017b,SPeng18_PVNet,YuXiang17_PoseCNN}, we measure the accuracy of pose estimation using the \textit{ADD score} \cite{hinterstoisserModelBasedTraining2013}. ADD is defined as an average Euclidean distance between model vertices transformed with the predicted and the ground truth pose. More formally it is defined as follows:

\begin{equation}
\label{add_standard}
m = \avgunder_{\mathbf{x} \in \mathcal{M}} \norm{ (\mathbf{Rx} + \mathbf{t}) - (\mathbf{\hat{R}x} + \mathbf{\hat{t}}) }_2,
\end{equation}
where $\mathcal{M}$ is a set of vertices of a particular model, $\mathbf{R}$ and $\mathbf{t}$ are the rotation and translation of a ground truth transformation whereas $\mathbf{\hat{R}}$ and $\mathbf{\hat{t}}$ correspond to those of an estimated transformation. The ADD metric can be extended in order to handle symmetric objects as in \cite{hinterstoisserModelBasedTraining2013}:
\begin{equation}
\label{add_symmetric}
m = \avgunder_{\mathbf{x}_2 \in \mathcal{M}} \minunder_{\mathbf{x}_1 \in \mathcal{M}} \norm{ (\mathbf{Rx}_1 + \mathbf{t}) - (\mathbf{\hat{R}x}_2 + \mathbf{\hat{t}}) }_2
\end{equation}
Instead of measuring distance from a predicted location of each particular model's vertex to its ground truth location, it suggests to take the closest vertex of the model transformed with the ground truth transformation.

Conventionally, a pose is considered correct if ADD is smaller than the $10\%$ of the model's diameter. The accuracy of pose estimation is reported as the percentage of correctly estimated poses.


\subsection{Single Object Pose Estimation}

Results of the pose estimation experiments on the LineMOD dataset are reported in Table \ref{tab:add}. We separately compared our method trained either on real data or on synthetic data. The table provides the comparison of deep learning-based refinement pipelines as well. The left-hand side of the table reports the accuracy of pose estimation as percentages of poses which are correct according to the ADD measure for the training done on synthetic data. If no refinement is used, our approach outperforms all other approaches by a significant margin on the majority of the objects. Moreover, the average percentage of correctly estimated poses ($50\%$) is significantly higher than $28.65\%$ of the second best approach. The accuracy gap is more prominent on small objects such as the ape and duck. The availability of a large number of 2D-3D correspondences ensures that the performance of our method is $5$ times better than SSD6D's and almost $2$ times better than AAE's. If deep learning-based refinement is used, we significantly outperform \cite{manhardtDeepModelBased6D} with $66.43\%$ of correct poses against $34.1\%$.

\begin{figure}[t]
    \centering
    \includegraphics[width=\linewidth]{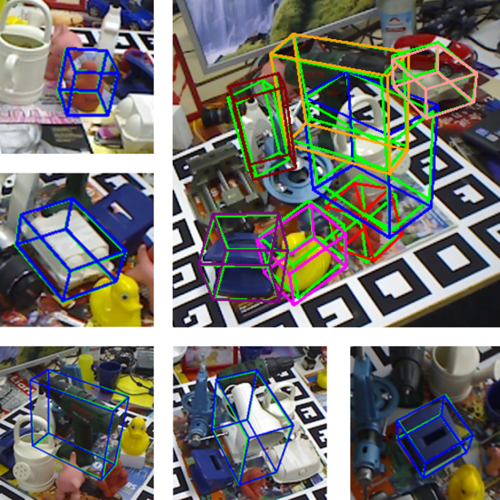} %
    \caption{\textbf{Qualitative results:} Poses predicted with the proposed approach on (a) the LineMOD dataset and (b) the OCCLUSION dataset. Green bounding boxes correspond to ground truth poses, bounding boxes of other colors to predicted poses. For both datasets predicted poses are very close to correct poses.\label{fig:linemod_examples}}%
    \vspace{-1em}
\end{figure}

If trained on real data, our method is the second best after \cite{SPeng18_PVNet}. The right-hand side of Table \ref{tab:add} compares the proposed approach to the previous deep learning-based ones. If no refinement is used, the proposed approach outperforms PoseCNN and YOLO6D by a significant margin, while performing on par with PVNet on most of the objects. On average, we are better than PoseCNN by $31\%$, YOLO6D by $23.57\%$. Again, our approach uses RGB data exclusively and does not rely on depth data. Figure \ref{fig:linemod_examples} provides a visual comparison of ground truth poses versus predicted poses. Poses are visualized as projections of 3D bounding boxes of models in given poses on top of a test image. In comparison to deep learning-based refinement of \cite{LiWJXF18_DeepIM}, we perform on average better by $6.55\%$ reaching $95.15\%$ of correct poses. When DeepIM was applied to the poses predicted by the proposed approach, ADD improved to $91.8\%$ which is better than the original $88.6\%$ reported in their paper, but still worse than the result of our refiner.

\setlength{\tabcolsep}{3pt}
\begin{table}[b]
 \vspace{-1em}
  \centering
  \caption{\textbf{Pose estimation for multiple objects:} Comparison of our approach on real data to the other RGB detectors on the OCCLUSSION dataset. The table reports percentages of correctly estimated poses w.r.t. the ADD score.}
  \resizebox{1\linewidth}{!}{
    \begin{tabular}{r|ccccccc}
    \toprule
    \textbf{Method}& 
    \parbox{1.4cm}{\centering \textbf{YOLO6D}\\ \cite{tekinRealTimeSeamlessSingle2017}} & 
    \parbox{1.3cm}{\centering \textbf{PoseCNN} \\ \cite{YuXiang17_PoseCNN}} &  
    \parbox{1.48cm}{\centering \textbf{SSD6D} \\ \textbf{+ Ref} \cite{manhardtDeepModelBased6D} }  & 
    \parbox{1.1cm}{\centering \textbf{HMap} \\ \cite{Oberweger_2018_ECCV}} & 
    \parbox{1.1cm}{\centering \textbf{PVNet} \\ \cite{SPeng18_PVNet}} & 
    \parbox{0.7cm}{\centering \vspace{0.3em} \textbf{Ours} \vspace{0.3em}} & 
    \parbox{1.0cm}{\centering \textbf{Ours} \\ \textbf{+Ref}} \\
    \midrule
    \textbf{Mean} & 6.42 & 24.9 & 27.5 & 30.4 & 40.77 & 32.79 & \textbf{47.25} \\
    \bottomrule
    \end{tabular}%
    }
  \label{tab:occlusion_add}%

\end{table}%

In conclusion, the proposed detector achieves state-of-the-art results surpassing other detectors by a large margin on synthetic data and performs either much better or comparable to the other detectors on real data. The proposed refinement clearly outperforms all the competitors both on real and synthetic data. Pose quality varies from object to object, but in general poses are significantly better for larger objects since there are more 2D-3D correspondences available. On the other hand, simplicity of the proposed approach also makes it quick. On average our detector performs at 33 FPS. The runtime can be adjusted by changing the number of RANSAC iterations, as it is the bottleneck of the pipeline. One iteration of the refinement takes 5ms, excluding the rendering time, which heavily depends on the renderer used. Two refinement iterations suffice for synthetic data, one iteration---for real data.

\setlength{\tabcolsep}{6pt}
\begin{table}[t]
  \centering
  \caption{\textbf{Detection performance for multiple objects:} Comparison of the state-of-the-art mean average precision (mAP) scores on the OCCLUSION dataset.}
  \resizebox{1\columnwidth}{!}{
    \begin{tabular}{r|cccc}
    \toprule
    \textbf{Method} & \textbf{SSD6D} \cite{kehlSSD6DMakingRGBbased2017b}  & \textbf{YOLO6D} \cite{tekinRealTimeSeamlessSingle2017}  & \textbf{Brachmann} \cite{brachmannUncertaintyDriven6DPose2016a} & \textbf{Ours} \\
    \midrule
    \textbf{mAP} & 0.38 & 0.48 & 0.51 & 0.48 \\
    \bottomrule
    \end{tabular}%
    }
  \label{tab:map}%
  \vspace{-1em}
\end{table}%

\subsection{Multiple Object Pose Estimation}

Performance evaluation of the proposed detector in cases when the number of objects to detect increases and when severe occlusions are present was conducted on the OCCLUSION dataset \cite{brachmannLearning6DObject2014a}. Accuracy of object detection on the OCCLUSION dataset is conventionally reported in terms of mean average precision (mAP). 
The confidence score is computed based on the RANSAC inlier proportion as confidence, rendering the final score of 0.48, which is comparable the best result on this dataset (see Table \ref{tab:map}). Table \ref{tab:occlusion_add} demonstrates ADD scores for various detectors on the OCCLUSION dataset. Before the refinement, the proposed detector shows very competitive results in comparison to other detectors. After the refinement, the proposed approach performs substantially better and achieves the best results.

\section{Conclusion}

In this paper we proposed the Dense Pose Object Detector (DPOD) method that regresses multi-class object masks and dense 2D-3D correspondences between image pixels and corresponding 3D models. Unlike the best performing methods that regress projections of the object's bounding boxes \cite{radBB8ScalableAccurate2017b, tekinRealTimeSeamlessSingle2017} or formulate pose estimation as a discrete pose classification problem \cite{kehlSSD6DMakingRGBbased2017b}, dense correspondences computed by our method allow for more robust and accurate 6D pose estimation. We demonstrated that for both, real and synthetic training data, our detector outperforms other related works, such as \cite{tekinRealTimeSeamlessSingle2017,YuXiang17_PoseCNN}, by a large margin and performs similarly to \cite{SPeng18_PVNet}. The proposed pose refinement approach also performs very well and allows for achieving a pose accuracy that surpasses all other related deep learning-based pose refinement approaches, while having a simpler and more lightweight backbone architecture. 
\appendix
\section{Supplementary Material}
\subsection{Implementation Details}

\begin{figure}[tbp]
	\centering
	\includegraphics[width=1\linewidth]{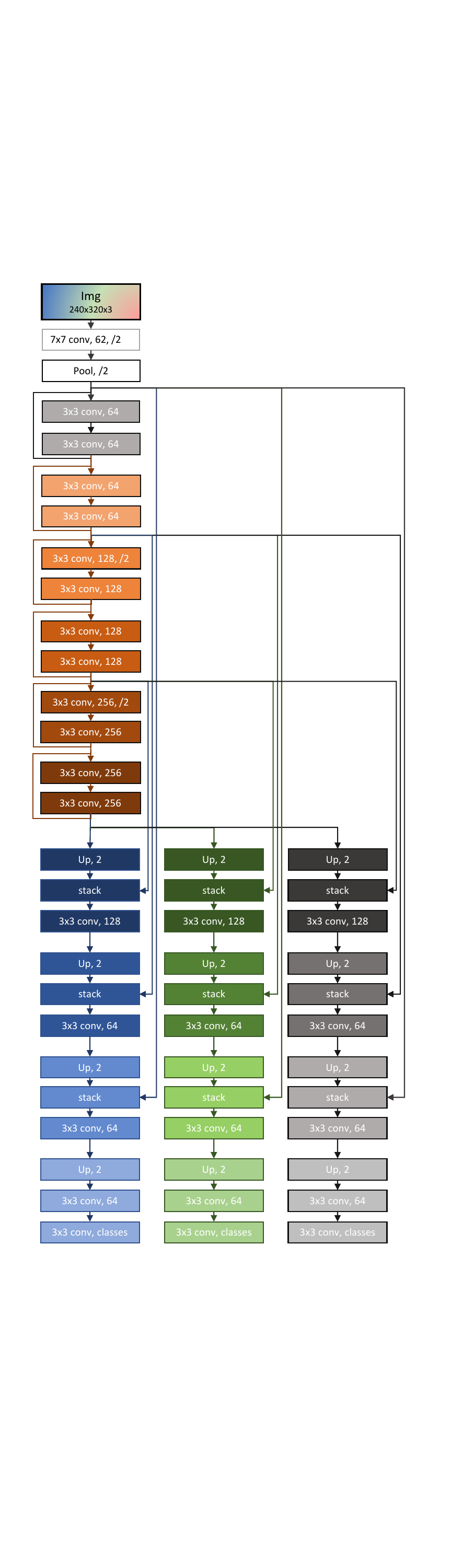}
	\caption{\textbf{DPOD's network architecture:} Encoder-decoder architecture based on ResNet.}
	\label{figure:net}
	\vspace{-1.2cm}
\end{figure}

The architecture of our detector is visualized in Figure~\ref{figure:net}. The refinement network utilizes the same backbone architecture. It is a standard ResNet-like (ResNet18 in PyTorch) model with a reduced number of layers and pooling operations in comparison to the original ResNet first presented in \cite{heDeepResidualLearning2016}. Upsampling is implemented as bilinear interpolation rather than deconvolution in order to decrease the number of parameters and the required amount of computations. Each upsampling is followed by the concatenatination of the output feature map with the feature map from the previous level, and one convolutional layer. When the detector is trained on synthetic data, the first five layers are frozen in order to prevent overfitting to peculiarities of the rendered data. The architecture of the refinement network follows the same architectural idea, except for the absence of upsampling and presence of fully-connected layers at the end. Again, the first five layers are used in siamese-like fashion for extracting features from image crops and renderings.

\subsection{RANSAC Iterations}
The number of RANSAC iterations crucially influences the quality of predicted poses. We ended up using 150 iterations as it yielded the best trade off between quality and runtime. The larger amount of iterations generally did not improve the results significantly, but resulted in longer execution times (see Table~\ref{tab:ransac}). Additionally, the ADD scores after one iteration of the proposed refinement are provided. They show that even 25 iterations of RANSAC are enough to beat the state-of-the-art results if the refinement is used. More iterations of RANSAC do not result in the considerable increase of pose quality.

\setlength{\tabcolsep}{2.5pt}
\begin{table}[htbp]
  \centering
  \caption{\textbf{RANSAC iterations test:} The effect of the number of RANSAC iterations on the overall ADD score.}
  \resizebox{1\columnwidth}{!}{%
    \begin{tabular}{r|ccccccccc}
    \toprule
    \multicolumn{1}{c|}{\textbf{RANSAC \#}} & \textbf{5} & \textbf{25} & \textbf{50} & \textbf{100} & \textbf{150} & \textbf{200} & \textbf{250} & \textbf{350} & \textbf{500} \\
    \midrule
    \textbf{ADD w/o ref} & 59.15 & 76.95 & 80.15 & 82.12 & 82.98 & 83.44 & 83.79 & 84.33 & 84.66 \\
    \textbf{ADD w/ ref} & 80.45 & 92.59 & 93.88 & 94.79 & 95.15 & 95.31 & 95.39 & 95.38 & 95.39 \\
    \midrule
    \textbf{RANSAC ms} & 2     & 6     & 10    & 17    & 23    & 28    & 33    & 42    & 54 \\
    \bottomrule
    \end{tabular}%
    }
  \label{tab:ransac}%
\end{table}%

\subsection{Runtime analysis}

\begin{table*}[t]
  \centering
  \caption{\textbf{Comparison of deep learning-based refinement methods:} Our refinement approach shows the overall best ADD score with respect to the latest state-of-the art method DeepIM~\cite{LiWJXF18_DeepIM}.}
  \resizebox{1\textwidth}{!}{%
    \begin{tabular}{r|ccccccccccccc|c}
    \toprule
    \textbf{Method/Object} & \textbf{Ape} & \textbf{Bench.} & \textbf{Cam} & \textbf{Can} & \textbf{Cat} & \textbf{Dril.} & \textbf{Duck} & \textbf{Eggb.} & \textbf{Gl.} & \textbf{Hol.} & \textbf{Iron} & \textbf{Lamp} & \textbf{Ph.} & \textbf{Avg.} \\
    \midrule
    \textbf{PoseCNN \cite{YuXiang17_PoseCNN} + DeepIM \cite{LiWJXF18_DeepIM}} & 77.0  & 97.5  & 93.5  & 96.5  & 82.1  & 95.0  & 77.7  & 97.1  & 99.4  & 52.8  & 98.3  & 97.5  & 87.7  & 88.6 \\
    \textbf{Ours + DeepIM \cite{LiWJXF18_DeepIM}} & 78.70 & 98.43 & \textbf{97.75} & 97.57 & 85.16 & 91.55 & 80.24 & 99.68 & \textbf{99.48} & 75.66 & 99.74 & \textbf{98.20} & 91.38 & 91.81 \\
    \textbf{Ours + Our ref.} & \textbf{87.73} & \textbf{98.45} & 96.07 & \textbf{99.71} & \textbf{94.71} & \textbf{98.8} & \textbf{86.29} & \textbf{99.91} & 96.82 & \textbf{86.87} & \textbf{100} & 96.84 & \textbf{94.69} & \textbf{95.15} \\
    \bottomrule
    \end{tabular}%
    }
  \label{tab:refinement}%
\end{table*}%

In Table~\ref{tab:runtimes} we provide the runtimes of the proposed approach for all models of the LineMOD dataset. The total runtime consists of the time needed for PnP and approximately $13$ ms for all the auxiliary tasks: the network's forward pass, post-processing of predicted segmentation, and computation of 2D-3D correspondences. Table \ref{tab:times} provides comparison of the runtime of our detector with all the main competitors mentioned in the paper.  All the experiments were conducted on an Intel Core i7-6900K CPU 3.20GHz with NVIDIA TITAN X (Pascal) GPU.

\setlength{\tabcolsep}{10pt}
\begin{table}[t]
  \centering
  \caption{\textbf{Runtime comparison:} Time-efficiency of our approach with respect to the other state-of-the-art approaches. \label{tab:times}}
  \resizebox{1\columnwidth}{!}{%
    \begin{tabular}{r|cc}
    \toprule
    \multicolumn{1}{c|}{\textbf{Method}} & \textbf{Frames per second} & \textbf{Refinement} \\
    \midrule
    \textbf{AAE \cite{sundermeyerImplicit3DOrientation2018}} & 4     & 200 ms/object \\
    \textbf{SSD6D \cite{kehlSSD6DMakingRGBbased2017b}} & 10    &  24 ms/object \\
    \textbf{PVNet \cite{SPeng18_PVNet}} & 25    & - \\
    \textbf{Ours} & 33    & 5 ms/object \\
    \textbf{YOLO6D \cite{tekinRealTimeSeamlessSingle2017}} & 50    & - \\
    \bottomrule
    \end{tabular}%
    }
\end{table}%

\subsection{Refinement}

\setlength{\tabcolsep}{9pt}
\begin{table}[b]
  \centering
  \caption{\textbf{Runtime analysis:} Runtime of the proposed approach for all models of the LineMOD dataset.}
  \resizebox{1\columnwidth}{!}{
    \begin{tabular}{r|cc|c}
    \toprule
    \textbf{Model} & \textbf{PnP + RANSAC (ms)} & \textbf{Total (ms)} & \textbf{FPS} \\
    \midrule
    Ape & 7     & 20    & 50 \\
    Benchvise & 40    & 51    & 20 \\
    Cam & 35    & 49    & 20 \\
    Can & 30    & 44    & 23 \\
    Cat & 20    & 33    & 30 \\
    Driller & 26    & 40    & 25 \\
    Duck & 4     & 16    & 63 \\
    Eggbox & 9     & 23    & 43 \\
    Glue & 5     & 17    & 59 \\
    Holepuncher & 20    & 31    & 32 \\
    Iron & 34    & 48    & 21 \\
    Lamp & 40    & 54    & 19 \\
    Phone & 31    & 45    & 22 \\
    \midrule
    \textbf{Average} & 23    & 36    & 33 \\
    \bottomrule
    \end{tabular}%
    }
  \label{tab:runtimes}
\end{table}%

	
	

DeepIM~\cite{LiWJXF18_DeepIM} presents an iterative refinement routine that takes an initial pose estimate from any external detector and iteratively improves it. An additional per-model evaluation is provided (see Table \ref{tab:refinement}) to have a fair comparison of DeepIM with our pose refinement. It compares the following ADD scores: 1) ADD reported in the original DeepIM paper \cite{LiWJXF18_DeepIM}, which used PoseCNN \cite{YuXiang17_PoseCNN} to predict initial poses, 2) ADD if DeepIM is applied to poses predicted by our detector, 3) ADD if poses predicted by the proposed detector are refined with the proposed refinement. It is important to mention that two iterations of DeepIM were made, as was suggested in the paper. The proposed refinement was run only for one iteration. The table clearly shows that better initial pose hypotheses allow for better results after refinement. It is also clear that our refinement clearly outperforms DeepIM on most of the objects, while performing only insignificantly worse on others.  

\subsection{Correspondence Quality}
In this section, we demonstrate the quality of the output correspondences. Namely, each classified correspondence point is mapped to 3D and compared to the ground truth 3D point. The ground truth 3D points are obtained in exactly the same way as predicted points, i.e., by matching a UV map rendered in the ground truth pose to model's vertices.

\begin{figure}[b]
    \centering
    \includegraphics[width=1\columnwidth]{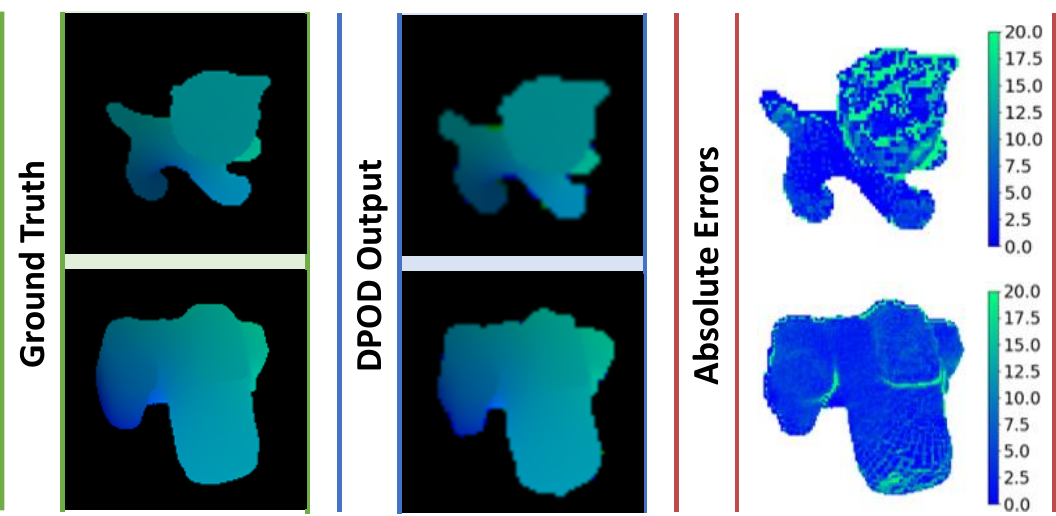}
    \caption{\textbf{Qualitative correspondence quality:} Comparison of ground truth (left), predicted (center) UV maps and heat maps (right) of absolute errors.}%
    \label{fig:correspondences}
\end{figure}

The results per object are shown in Table \ref{tab:correspondences}. The table reports the quality of correspondences separately for real and synthetic data. For each model, mean absolute error, median absolute error, and standard deviation of absolute errors are reported in millimeters. Relatively large mean error is explained by outliers, some of which can be quite significant. Therefore, median is a better measure due to its robustness to outliers. The table shows that the median error is consistent across all the models. Additionally, it demonstrates that the median error for the detector trained on real data is noticeably lower than for the detector trained on synthetic data. This explains the superior performance of training on real data.

\setlength{\tabcolsep}{6pt}
\begin{table}[t]
\caption{\textbf{Quantative correspondence quality:} Correspondence quality for real and synthetic data estimated in terms of mean and median absolute errors, and standard deviation.}
\centering
\resizebox{1\columnwidth}{!}{
\begin{tabular}{r|ccc|ccc}
\toprule
        & \multicolumn{3}{c|}{\textbf{Real Data}} & \multicolumn{3}{c}{\textbf{Synt Data}} \\
        \midrule
\textbf{Model}   & \textbf{Mean}     & \textbf{Median}   & \textbf{Std}     & \textbf{Mean}     & \textbf{Median}   & \textbf{Std}     \\
\midrule
Ape     & 10.05    & 4.58     & 14.60   & 11.46    & 5.74     & 15.26   \\
Benc.   & 10.36    & 4.70     & 19.29   & 15.92    & 6.99     & 25.71   \\
Cam     & 6.57     & 4.58     & 10.11   & 13.31    & 7.23     & 20.23   \\
Can     & 8.19     & 4.03    & 13.46   & 11.97    & 5.10     & 18.72   \\
Cat     & 8.60     & 4.77     & 12.22   & 9.87     & 5.42     & 13.99   \\
Driller & 8.52     & 4.78     & 17.78   & 18.14    & 6.80     & 36.06   \\
Duck    & 5.93     & 3.98     & 8.72    & 7.63     & 4.99     & 10.41   \\
Eggbox  & 6.00     & 4.26     & 10.23   & 42.39    & 9.40     & 48.07   \\
Glue    & 7.82     & 4.26     & 13.73   & 17.12    & 8.11     & 23.19   \\
Holep.  & 8.25     & 4.87     & 13.30   & 11.28    & 6.81     & 16.04   \\
Iron    & 7.18     & 4.51     & 12.31   & 11.06    & 6.89     & 17.34   \\
Lamp    & 11.64    & 4.31     & 24.80   & 18.60    & 8.58     & 30.85   \\
Phone   & 6.09     & 2.84     & 12.94   & 9.52     & 4.38     & 18.31  \\
\bottomrule
\end{tabular}}
\label{tab:correspondences}
\end{table}

Figure~\ref{fig:correspondences} provides a visual comparison of predicted and ground truth UV maps and heat maps, which demonstrate where imprecisions take place. One can see that most imprecisions are concentrated on the outer boundaries of the object and, for objects with more complex geometry, on the edges of their structural elements, i.e. in places where rapid correspondence value changes occur.

\subsection{Multiple Instance Detection}
Our detector also works when multiple instances of the same object are presented, because we parse through the regions of the output mask when the network forward pass is complete. The only limitation comes into play when several objects of the same class overlap and form a single region. In this case, only one pose will be estimated instead of two.

\begin{figure}[b]
    \centering
    \includegraphics[width=1\columnwidth]{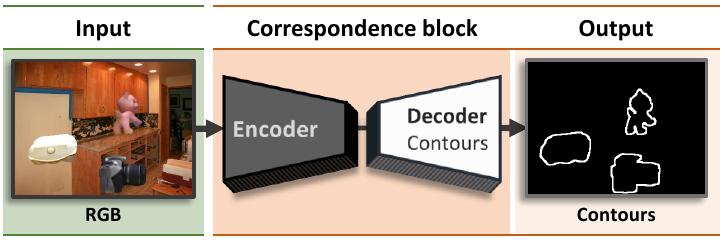}
    \caption{\textbf{Contour regression:} Additional contour regression head for multiple instance detection.}%
    \label{fig:contours}
\end{figure}

\begin{figure}[t]
    \centering
    \includegraphics[width=1\columnwidth]{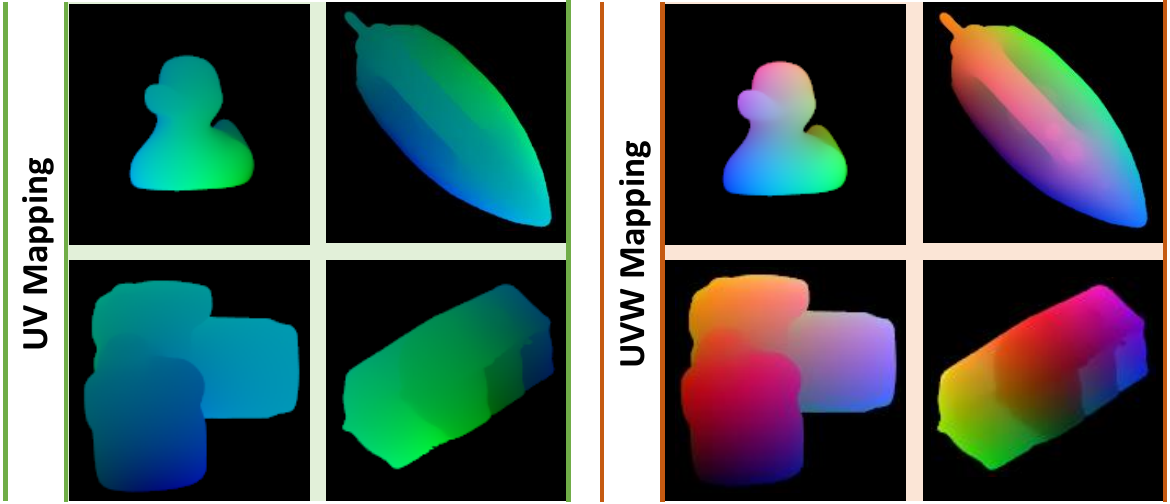}
    \caption{\textbf{UVW mapping:} Visual comparison between UV and UVW mappings.}%
    \label{fig:uvw}
\end{figure}
To overcome this, an additional contour regression head can be added to the correspondence block (see Figure~\ref{fig:contours}). Once regressed, the output contours are simply multiplied with the ID mask forming different regions, which, as a result, allows to distinguish between different regions of the same class.

\subsection{UVW Mapping}
While being computationally efficient, the UV mapping has a number of drawbacks. In the majority of cases, a simple spherical projection is sufficient to achieve a satisfactory quality of the mapping. However, certain cases can require a different treatment to minimize the stretching effect where one color can cover several model vertices due to discretization. This is especially a big problem for more complicated geometries, which in some cases might require a selection of another projection type or even a manual UV mapping for reaching the best results.

A straightforward solution to this is the UVW mapping based on normalized 3D coordinates of the model. Instead of 2-channel UV maps, we then have 3-channel UVW maps that are again discretized to the range [0, 255]. The only algorithmic adjustment that has to be done is an additional W-channel classification head. While decreasing the memory efficiency and increasing a computational complexity of the network (due to a higher-dimensional solution space, i.e., $256^3$ instead of $256^2$ in case of UV mapping), it has an advantage of providing better quality correspondences (especially for objects with complex geometries) and of being fully automatic (see Figure~\ref{fig:uvw} for visual comparison).

Our additional experimental ablations have shown an almost identical performance on the LineMOD and OCCLUSION datasets, but slightly higher execution times and memory requirements. Nevertheless, despite the increased complexity, we believe that this extension would prove itself useful in many real-world applications.

\subsection{Additional Qualitative Results}
\begin{figure}[t]
	\centering
	\begin{adjustbox}{minipage=\textwidth,scale=1}
	\subfloat{\includegraphics[width=0.16\textwidth]{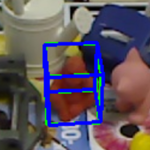}}
	\hfill
	\subfloat{\includegraphics[width=0.16\textwidth]{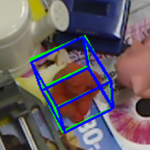}}
	\hfill
	\subfloat{\includegraphics[width=0.16\textwidth]{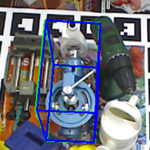}}
	\hfill
	\subfloat{\includegraphics[width=0.16\textwidth]{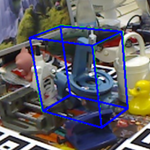}}
	\hfill
	\subfloat{\includegraphics[width=0.16\textwidth]{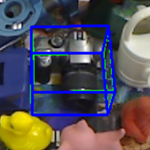}}
	\hfill
	\subfloat{\includegraphics[width=0.16\textwidth]{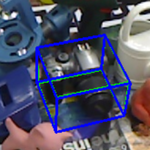}}
	\hfill
	
	\subfloat{\includegraphics[width=0.16\textwidth]{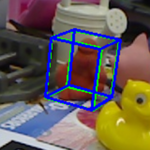}}
	\hfill
	\subfloat{\includegraphics[width=0.16\textwidth]{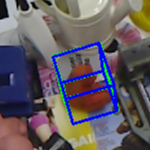}}
    \hfill
	\subfloat{\includegraphics[width=0.16\textwidth]{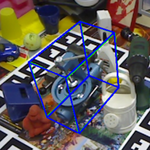}}
	\hfill
	\subfloat{\includegraphics[width=0.16\textwidth]{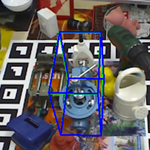}}
    \hfill
	\subfloat{\includegraphics[width=0.16\textwidth]{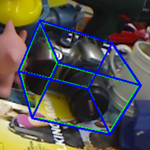}}
	\hfill
	\subfloat{\includegraphics[width=0.16\textwidth]{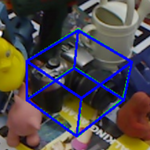}}
	\hfill

	\subfloat{\includegraphics[width=0.16\textwidth]{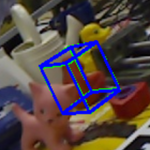}}
	\hfill
	\subfloat{\includegraphics[width=0.16\textwidth]{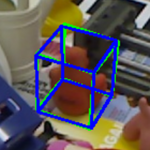}}
	\hfill
	\subfloat{\includegraphics[width=0.16\textwidth]{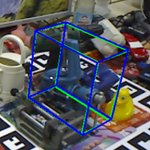}}
	\hfill
	\subfloat{\includegraphics[width=0.16\textwidth]{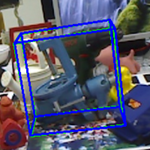}}
	\hfill
	\subfloat{\includegraphics[width=0.16\textwidth]{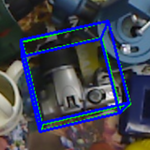}}
	\hfill
	\subfloat{\includegraphics[width=0.16\textwidth]{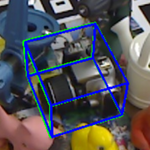}}
	\hfill
	
	\subfloat{\includegraphics[width=0.16\textwidth]{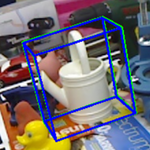}}
	\hfill
	\subfloat{\includegraphics[width=0.16\textwidth]{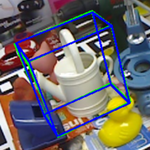}}
	\hfill
	\subfloat{\includegraphics[width=0.16\textwidth]{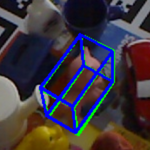}}
	\hfill
	\subfloat{\includegraphics[width=0.16\textwidth]{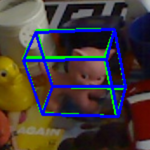}}
	\hfill
	\subfloat{\includegraphics[width=0.16\textwidth]{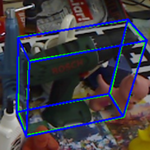}}
	\hfill
	\subfloat{\includegraphics[width=0.16\textwidth]{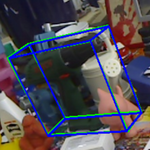}}
	\hfill

	\subfloat{\includegraphics[width=0.16\textwidth]{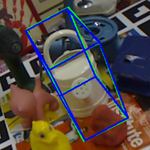}}
	\hfill
	\subfloat{\includegraphics[width=0.16\textwidth]{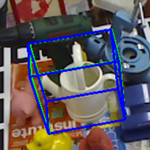}}
	\hfill
	\subfloat{\includegraphics[width=0.16\textwidth]{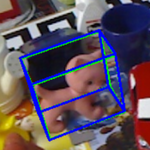}}
	\hfill
	\subfloat{\includegraphics[width=0.16\textwidth]{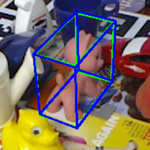}}
	\hfill
	\subfloat{\includegraphics[width=0.16\textwidth]{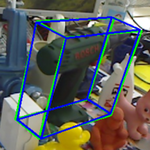}}
	\hfill
	\subfloat{\includegraphics[width=0.16\textwidth]{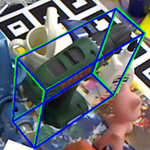}}
	\hfill
	
	\subfloat{\includegraphics[width=0.16\textwidth]{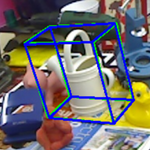}}
	\hfill
	\subfloat{\includegraphics[width=0.16\textwidth]{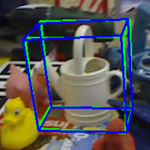}}
	\hfill
	\subfloat{\includegraphics[width=0.16\textwidth]{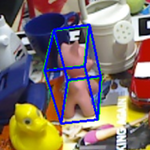}}
	\hfill
	\subfloat{\includegraphics[width=0.16\textwidth]{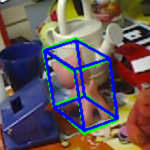}}
	\hfill
	\subfloat{\includegraphics[width=0.16\textwidth]{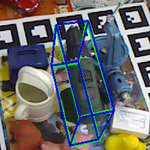}}
	\hfill
	\subfloat{\includegraphics[width=0.16\textwidth]{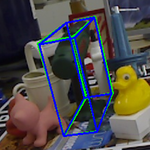}}
	\hfill
	
	\caption{\textbf{Example results on the LineMOD dataset:} ape, can (left), benchvise, cat (middle), cam, driller (right). Green bounding boxes correspond to ground truth poses, blue bounding boxes correspond to predicted poses.\label{figure:vis_LineMOD_1}}
	\end{adjustbox}
\end{figure}

\begin{figure*}[!tbp]
	\centering
	
	\subfloat{\includegraphics[width=0.16\textwidth]{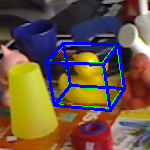}}
	\hfill
	\subfloat{\includegraphics[width=0.16\textwidth]{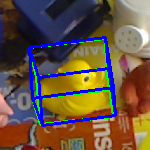}}
	\hfill
	\subfloat{\includegraphics[width=0.16\textwidth]{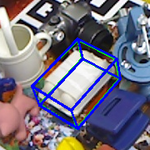}}
	\hfill
	\subfloat{\includegraphics[width=0.16\textwidth]{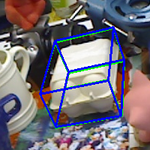}}
	\hfill
	\subfloat{\includegraphics[width=0.16\textwidth]{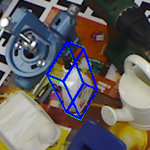}}
	\hfill
	\subfloat{\includegraphics[width=0.16\textwidth]{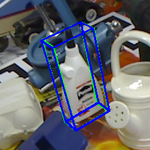}}
	\hfill

	\subfloat{\includegraphics[width=0.16\textwidth]{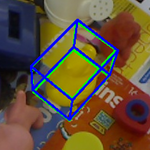}}
	\hfill
	\subfloat{\includegraphics[width=0.16\textwidth]{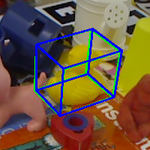}}
	\hfill
	\subfloat{\includegraphics[width=0.16\textwidth]{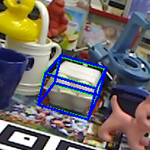}}
	\hfill
	\subfloat{\includegraphics[width=0.16\textwidth]{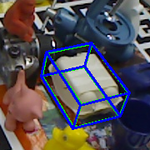}}
	\hfill
	\subfloat{\includegraphics[width=0.16\textwidth]{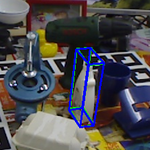}}
	\hfill
	\subfloat{\includegraphics[width=0.16\textwidth]{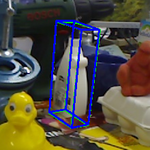}}
	\hfill
	
	\subfloat{\includegraphics[width=0.16\textwidth]{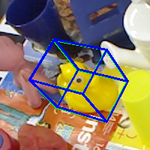}}
	\hfill
	\subfloat{\includegraphics[width=0.16\textwidth]{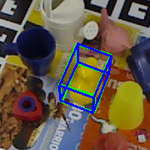}}
	\hfill
	\subfloat{\includegraphics[width=0.16\textwidth]{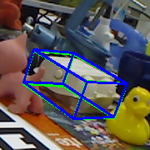}}
	\hfill
	\subfloat{\includegraphics[width=0.16\textwidth]{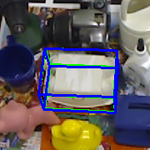}}
	\hfill
	\subfloat{\includegraphics[width=0.16\textwidth]{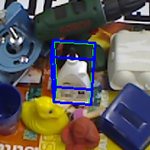}}
	\hfill
	\subfloat{\includegraphics[width=0.16\textwidth]{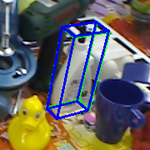}}
	\hfill
	
	\subfloat{\includegraphics[width=0.16\textwidth]{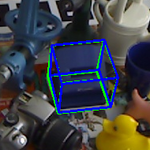}}
	\hfill
	\subfloat{\includegraphics[width=0.16\textwidth]{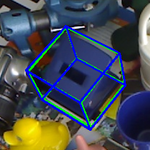}}
	\hfill
	\subfloat{\includegraphics[width=0.16\textwidth]{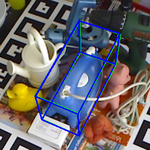}}
	\hfill
	\subfloat{\includegraphics[width=0.16\textwidth]{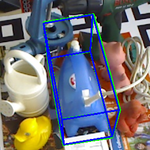}}
	\hfill
	\subfloat{\includegraphics[width=0.16\textwidth]{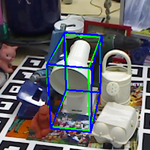}}
	\hfill
	\subfloat{\includegraphics[width=0.16\textwidth]{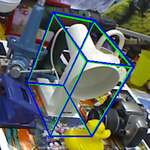}}
	\hfill

	\subfloat{\includegraphics[width=0.16\textwidth]{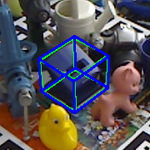}}
	\hfill
	\subfloat{\includegraphics[width=0.16\textwidth]{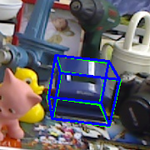}}
	\hfill
	\subfloat{\includegraphics[width=0.16\textwidth]{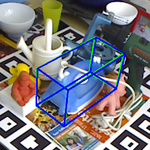}}
	\hfill
	\subfloat{\includegraphics[width=0.16\textwidth]{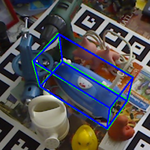}}
	\hfill
	\subfloat{\includegraphics[width=0.16\textwidth]{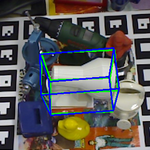}}
	\hfill
	\subfloat{\includegraphics[width=0.16\textwidth]{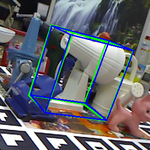}}
	\hfill
	
	\subfloat{\includegraphics[width=0.16\textwidth]{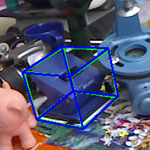}}
	\hfill
	\subfloat{\includegraphics[width=0.16\textwidth]{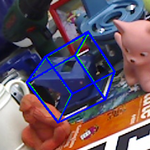}}
	\hfill
	\subfloat{\includegraphics[width=0.16\textwidth]{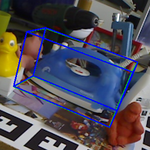}}
	\hfill
	\subfloat{\includegraphics[width=0.16\textwidth]{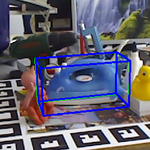}}
	\hfill
	\subfloat{\includegraphics[width=0.16\textwidth]{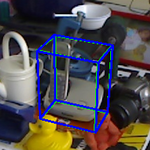}}
	\hfill
	\subfloat{\includegraphics[width=0.16\textwidth]{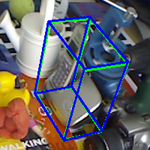}}
	\hfill

	\caption{\textbf{Example results on the LineMOD dataset:} duck, holepuncher (left), eggbox, iron (middle), glue, lamp, phone (right). Green bounding boxes correspond to ground truth poses, blue bounding boxes correspond to predicted poses.\label{figure:vis_LineMOD_2}}
\end{figure*}

\begin{figure}[!h]
	\centering
	\begin{adjustbox}{minipage=\textwidth,scale=1}
	
	\hfill
	\subfloat{\includegraphics[width=0.475\textwidth]{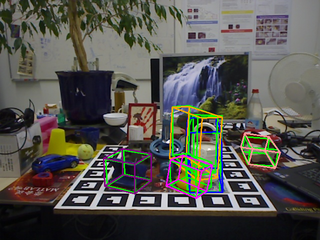}}
	\hfill
	\subfloat{\includegraphics[width=0.475\textwidth]{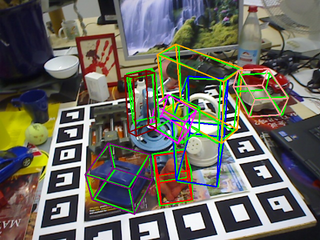}}
	\hfill
	
	\hfill
	\subfloat{\includegraphics[width=0.475\textwidth]{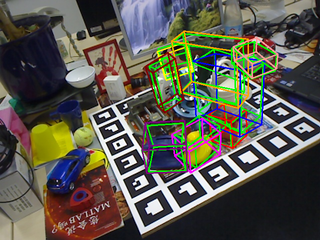}}
	\hfill
	\subfloat{\includegraphics[width=0.475\textwidth]{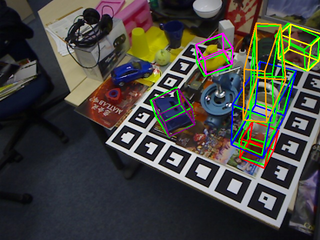}}
	\hfill
	
	\hfill
	\subfloat{\includegraphics[width=0.475\textwidth]{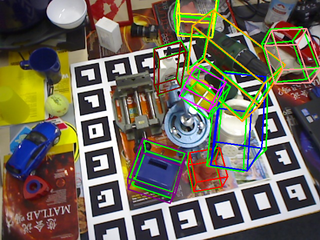}}
	\hfill
	\subfloat{\includegraphics[width=0.475\textwidth]{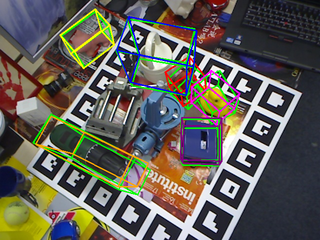}}
	\hfill
	
	\caption{\textbf{Example results on the OCCLUSION dataset.} Green bounding boxes correspond to ground truth poses, bounding boxes of other colors correspond to predicted poses.\label{figure:vis_occlusion1}}
	\end{adjustbox}
\end{figure}

\begin{figure*}[!tbp]
	\centering
	
	\hfill
	\subfloat{\includegraphics[width=0.475\textwidth]{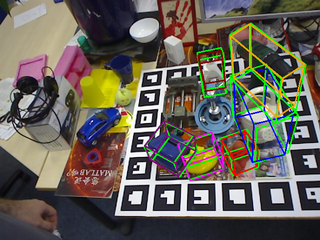}}
	\hfill
	\subfloat{\includegraphics[width=0.475\textwidth]{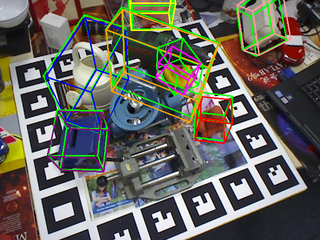}}
	\hfill
	
	\hfill
	\subfloat{\includegraphics[width=0.475\textwidth]{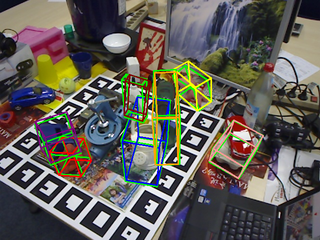}}
	\hfill
	\subfloat{\includegraphics[width=0.475\textwidth]{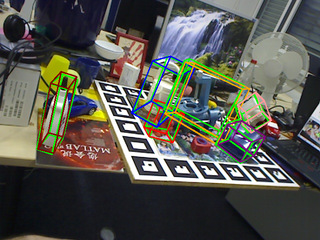}}
	\hfill
	
	\hfill
	\subfloat{\includegraphics[width=0.475\textwidth]{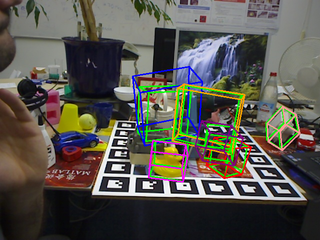}}
	\hfill
	\subfloat{\includegraphics[width=0.475\textwidth]{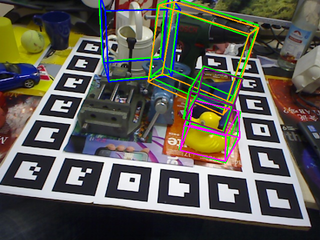}}
	\hfill
	
	\caption{\textbf{Example results on the OCCLUSION dataset.} Green bounding boxes correspond to ground truth poses, bounding boxes of other colors correspond to predicted poses.\label{figure:vis_occlusion2}}
\end{figure*}

In Figures~\ref{figure:vis_LineMOD_1} to \ref{figure:vis_occlusion2}, we show additional qualitative pose results on LineMOD and OCCLUSION datasets. Our method demonstrates very high quality poses and is robust to occlusions and illumination changes.


\clearpage
{\small
\bibliographystyle{ieee_fullname}
\bibliography{egbib}
}


\end{document}